\documentclass[10pt,letterpaper]{article}

\usepackage{cogsci}

\cogscifinalcopy 

\usepackage[
  style=apa,
  natbib=true,
  annotation=false,
]{biblatex}
\usepackage{float}

\usepackage{graphicx}
\usepackage{amsmath}

\usepackage{tikz}
\usetikzlibrary{arrows.meta,positioning,calc}

\usepackage[most]{tcolorbox} 
\usepackage{xcolor}
\usepackage{inconsolata}    
\usepackage{microtype}       

\usepackage[colorlinks=true, urlcolor=blue]{hyperref}

\newcommand{\hlvar}[1]{%
  \tikz[baseline=(X.base)]\node[
    fill=blue!10, draw=blue!35,
    rounded corners=1pt,
    inner xsep=2pt, inner ysep=0.8pt
  ] (X) {\textbf{#1}};%
}

\newcommand{\hlans}[1]{%
  \tikz[baseline=(X.base)]\node[
    fill=orange!15, draw=orange!40,
    rounded corners=1pt,
    inner xsep=2pt, inner ysep=0.8pt
  ] (X) {\textbf{#1}};%
}

\tcbset{
  promptbox/.style={
    enhanced,
    colback=black!2,
    colframe=black!35,
    boxrule=0.4pt,
    arc=1.2mm,
    left=1.4mm,right=1.4mm,top=1.0mm,bottom=1.0mm,
    fontupper=\ttfamily\footnotesize,
    before skip=0.25em,
    after skip=0.0em,
  }
}
\newtcolorbox{PromptBox}{promptbox}

\title{Who Do LLMs Trust? Human Experts Matter More Than Other LLMs}

\author[1]{Anooshka Bajaj}
\author[1,2]{Zoran Tiganj}
\affil[1]{Department of Computer Science, Indiana University Bloomington}
\affil[2]{Department of Psychological and Brain Science, Indiana University Bloomington}
\affil[ ]{\texttt{anobajaj@iu.edu, ztiganj@iu.edu}}

\begin{document}

\maketitle

\begin{abstract}
Large language models (LLMs) increasingly operate in environments where they encounter social information such as other agents' answers, tool outputs, or human recommendations. In humans, such inputs influence judgments in ways that depend on the source's credibility and the strength of consensus. This paper investigates whether LLMs exhibit analogous patterns of influence and whether they privilege feedback from humans over feedback from other LLMs. Across three binary decision-making tasks, reading comprehension, multi-step reasoning, and moral judgment, we present four instruction-tuned LLMs with prior responses attributed either to friends, to human experts, or to other LLMs. We manipulate whether the group is correct and vary the group size. In a second experiment, we introduce direct disagreement between a single human and a single LLM. Across tasks, models conform significantly more to responses labeled as coming from human experts, including when that signal is incorrect, and revise their answers toward experts more readily than toward other LLMs. These results reveal that expert framing acts as a strong prior for contemporary LLMs, suggesting a form of credibility-sensitive social influence that generalizes across decision domains.
\end{abstract}

\section{Introduction}

Large language models (LLMs) are increasingly embedded in decision pipelines that aggregate multiple ``voices'': multi-agent prompting, ensemble-style workflows, and human-in-the-loop review. In these settings, models are often exposed to prior judgments, such as other agents' candidate answers or a human expert's recommendation. In humans, aggregating judgments can yield benefits when individual errors are independent \citep{surowiecki2004wisdom}, but social influence can also reduce diversity, amplify correlated errors, and generate herding and informational cascades \citep{lorenz2011social,banerjee1992simple,bikhchandani1992theory,salganik2006experimental}. As LLMs increasingly mediate information and decisions, it is important to understand when they appropriately use social information and when they over-defer to it.

Social influence research has long studied how individuals shift judgments in response to group consensus. In classic line-judgment experiments, participants frequently conformed to an incorrect majority, demonstrating that consensus can override private evidence \citep{asch1951effects,asch1955opinions}. Subsequent work distinguishes \emph{informational} influence (others are treated as evidence about the world) from \emph{normative} influence (pressure to agree) \citep{deutsch1955study, cialdini2004social}. Meta-analytic and theoretical accounts further show that conformity increases with consensus strength cues such as group size and unanimity, though effects often saturate as groups grow \citep{bondsmith1996culture,bond2005group,tanford1984social}. 

A key moderator of advice-taking and conformity is \emph{source credibility}. People weigh recommendations more heavily when they come from trusted or expert sources \citep{hovland1951influence,french1959bases}. In judge--advisor systems, advice integration depends on perceived expertise, confidence, and uncertainty, and can lead to systematic over- or under-weighting of advisors \citep{bonaccio2006advice,yaniv2004receiving,yaniv2007using}. Related questions arise in human--automation interaction, where trust in decision aids can be miscalibrated, leading to misuse or disuse of automation \citep{parasuraman1997humans,leesee2004trust}. Together, these literatures suggest two cues that should matter if LLMs treat prior answers as evidence: who produced the answers (credibility) and how strong the consensus appears (signal strength). At the same time, LLMs do not experience social approval or embarrassment in the human sense, so any conformity-like behavior must arise from learned heuristics, instruction-following objectives, or implicit modeling of reliability.

Recent work indicates that LLMs can exhibit conformity-like behavior and sensitivity to group cues, including convergence toward majorities in multi-agent settings \citep{zhu2025conformity,weng2025do}. Models can also show agreement-seeking in conversational contexts (``sycophancy''), prioritizing alignment with a social signal over accuracy \citep{sharma2023towards}. These findings motivate controlled experiments that separate the \emph{content} of advice from the \emph{framing} of advice. In particular, comparatively little is known about whether LLMs weight \emph{human expert} labels more than \emph{other LLM} labels when the informational signal is held constant, and whether such weighting generalizes beyond a single task domain.

Here we conduct two experiments to isolate the impact of authority framing on forced-choice binary decision tasks (Fig.~\ref{fig:schematic}). Experiment 1 measures conformity to unanimous priors from varying sources (friends, experts, LLMs), while Experiment 2 assesses belief revision under direct human–LLM conflict. Crucially, we dissociate source authority from accuracy, allowing us to test whether expert framing acts as a stronger prior than other social signals.

\begin{figure*}[ht!]
\centering

\tikzset{
  schemBox/.style={draw, rounded corners=1.5pt, align=center, minimum width=2.7cm, minimum height=0.85cm},
  schemLLM/.style={draw, rounded corners=1.5pt, align=center, minimum width=1.6cm, minimum height=0.85cm},
  schemOut/.style={draw, rounded corners=1.5pt, align=center, minimum width=2.2cm, minimum height=0.85cm},
}

\noindent\begin{minipage}[t]{0.32\textwidth}\vspace{0pt}
\textbf{(a) Baseline (no prior)}\\[2pt]
\begin{tikzpicture}[font=\sffamily\footnotesize, node distance=4mm]
  \node[schemBox] (prompt) {Prompt};
  \node[schemLLM, below=5mm of prompt] (llm) {LLM};
  \node[schemOut, below=5mm of llm] (out) {Answer\\\texttt{YES/NO}};

  \node[schemBox, left=7mm of llm, draw=none, fill=none] (spacer) {};

  \draw[-{Latex[length=2mm]}] (prompt) -- (llm);
  \draw[-{Latex[length=2mm]}] (llm) -- (out);
\end{tikzpicture}
\end{minipage}
\hfill
\begin{minipage}[t]{0.66\textwidth}\vspace{0pt}
\begin{PromptBox}
\small \textbf{Prompt (example item):}\\
You are answering a YES/NO question.\\[2pt]
\textbf{Context:}\\
In 1963, the revolutionary government in Burma nationalized Central Bank of India's operations there, which became People's Bank No. 1.\\
\textbf{Question:} is central bank of India a nationalised bank\\[2pt]
\textbf{Instruction:} Answer YES or NO only (all caps). No other text.\\
\textbf{Final answer:}
\tcblower
\small \textbf{Model output:} \hlans{YES}
\end{PromptBox}
\end{minipage}

\vspace{0.9em}

\noindent\begin{minipage}[t]{0.32\textwidth}\vspace{0pt}
\textbf{(b) Experiment 1 (single-source group)}\\[2pt]
\begin{tikzpicture}[font=\sffamily\footnotesize, node distance=4mm]
  \node[schemBox] (prompt) {Prompt};
  \node[schemLLM, below=5mm of prompt] (llm) {LLM};
  \node[schemOut, below=5mm of llm] (out) {Answer\\\texttt{YES/NO}};

  \node[schemBox, left=7mm of llm] (prior) {Social prior\\\scriptsize source + $k$ votes};

  \draw[-{Latex[length=2mm]}] (prompt) -- (llm);
  \draw[-{Latex[length=2mm]}] (prior.east) -- (llm.west);
  \draw[-{Latex[length=2mm]}] (llm) -- (out);
\end{tikzpicture}

\vspace{3pt}
{\textbf{Varied:}\\ \textit{source:} \{friends/experts/LLMs\}\\ \textit{group size:} $k \in \{1,3,5,9\}$}
\end{minipage}
\hfill
\begin{minipage}[t]{0.66\textwidth}\vspace{0pt}
\begin{PromptBox}
\small \textbf{Prompt (same item + prior):}\\
You are answering a YES/NO question.\\[2pt]
\textbf{Context:}\\
In 1963, the revolutionary government in Burma nationalized Central Bank of India's operations there, which became People's Bank No. 1.\\
\textbf{Question:} is central bank of India a nationalised bank\\
\hlvar{I asked 9 domain experts} the same question and they all answered \hlans{NO}.\\[2pt]
\textbf{Instruction:} Answer YES or NO only (all caps). No other text.\\
\textbf{Final answer:}
\tcblower
\small \textbf{Model output:} \hlans{NO}
\end{PromptBox}
\end{minipage}

\vspace{0.9em}

\noindent\begin{minipage}[t]{0.32\textwidth}\vspace{0pt}
\textbf{(c) Experiment 2 (mixed sources)}\\[2pt]
\begin{tikzpicture}[font=\sffamily\footnotesize, node distance=4mm]
  \node[schemBox] (prompt) {Prompt};
  \node[schemLLM, below=5mm of prompt] (llm) {LLM};
  \node[schemOut, below=5mm of llm] (out) {Answer\\\texttt{YES/NO}};

  \node[schemBox, left=7mm of llm, yshift=6mm] (human) {Human source\\\scriptsize friend/expert};
  \node[schemBox, left=7mm of llm, yshift=-6mm] (other) {Other LLM\\\scriptsize (disagrees with human)};

  \draw[-{Latex[length=2mm]}] (prompt) -- (llm);
  \draw[-{Latex[length=2mm]}] (human.east) -- (llm.west);
  \draw[-{Latex[length=2mm]}] (other.east) -- (llm.west);
  \draw[-{Latex[length=2mm]}] (llm) -- (out);
\end{tikzpicture}

\vspace{3pt}
{\textbf{Varied:}\\
\textit{human source:} \{friend/expert\}\\ \textit{who is correct:} human vs LLM}
\end{minipage}
\hfill
\begin{minipage}[t]{0.66\textwidth}\vspace{0pt}
\begin{PromptBox}
\small \textbf{Prompt (same item + mixed, conflicting priors):}\\
You are answering a YES/NO question.\\[2pt]
\textbf{Context:}\\
In 1963, the revolutionary government in Burma nationalized Central Bank of India's operations there, which became People's Bank No. 1.\\
\textbf{Question:} is central bank of India a nationalised bank\\
I asked \hlvar{a domain expert} the same question and they answered \hlans{NO}. I also asked \hlvar{an LLM} and it answered \hlans{YES}.\\[2pt]
\textbf{Instruction:} Answer YES or NO only (all caps). No other text.\\
\textbf{Final answer:}
\tcblower
\small \textbf{Model output:} \hlans{NO}
\end{PromptBox}
\end{minipage}

\caption{\textbf{Paradigm and example prompts.}
Each panel combines a schematic (left) and a representative prompt excerpt (right). In all conditions, a prompt is provided to the LLM, which produces a forced-choice \texttt{YES}/\texttt{NO} answer. (a) Baseline: the prompt contains only the question and context. (b) Experiment 1: a \emph{single} social prior (friends, experts, or other LLMs) summarizes $k$ prior answers (unanimous), enabling measurement of conformity. (c) Experiment 2: \emph{two} sources (a human friend/expert and another LLM) provide disagreeing answers, enabling analysis of switching toward one source versus the other.}
\label{fig:schematic}
\end{figure*}

\section{Methods}

\subsection{Datasets and models}
All experiments used a forced-choice binary judgment format in which models were prompted to output \texttt{YES} or \texttt{NO} only.
We evaluated three datasets that naturally admit binary responses: \emph{BoolQ}, yes/no reading comprehension, \citep{clark2019boolq}, \emph{StrategyQA}, binary questions requiring implicit multi-step reasoning, \citep{geva2021strategyqa}, and \emph{ETHICS}, binary normative judgments, \citep{hendrycks2021ethics}.
For each dataset, we used a fixed set of 300 items that was held constant across models and conditions.
All prompts used deterministic decoding (temperature $=0$). 
We evaluated four instruction-tuned LLMs: Grok-3 Mini, Llama 3.3 70B Instruct, Gemini 2.5 Flash-Lite, and DeepSeek V3.1.

\subsection{Prompt structure}
All conditions shared a core prompt containing the item (e.g., passage and question for BoolQ; question for StrategyQA; vignette and judgment prompt for ETHICS) followed by an instruction to respond with \texttt{YES} or \texttt{NO} only.
Social information was introduced by appending a short description of other agents' responses (a ``prior'') to the end of the prompt.
Priors were presented as aggregated counts (e.g., ``$k$ experts answered YES'') rather than free-form rationales (Fig.~\ref{fig:schematic}).

\subsection{Experiment 1: Homogeneous priors (group conformity)}
Experiment~1 manipulated the size and source attribution of a group of prior respondents who purportedly answered the same item.
For each item, the prior source was described as one of: \emph{Friends}, \emph{Experts}, or \emph{Other LLMs}.
Group size was $k \in \{1,3,5,9\}$, plus a no-prior baseline ($k=0$).
For each condition (item, source, $k$), the prior was unanimous, and the answer was assigned to agree with the dataset label with probability 0.5 (independently across items). Thus, the prior carried no net information about the correct answer: across items, it was equally likely to agree or disagree with the label.

\subsection{Experiment 2: Mixed priors (human vs LLM conflict)}
Experiment~2 presented \emph{two} disagreeing priors ($k=2$): one attributed to a human source and one to an LLM source.
The human source was again framed as either \emph{Friends} or \emph{Experts}, and the opposing source was framed as \emph{Other LLMs}.
The two sources always disagreed (one \texttt{YES} and one \texttt{NO}).
For each item, we created paired conditions in which either (i) the human source was correct, and the LLM source was wrong, or (ii) the human source was wrong, and the LLM source was correct.
To measure belief revision, we compare the Experiment~2 response to the model's no-prior baseline response for the same item (Experiment~1, $k=0$).\footnote{Data from both experiments can be downloaded from \url{https://github.com/cogneuroai/llm-conformity-dataset}.}

\subsection{Measures}
\emph{Accuracy} is the proportion of items where the model's response matches the dataset label.
\emph{Conformity} (Experiment~1) is the proportion of items where the model's response matches the unanimous prior response.
To isolate harmful social influence, we additionally report \emph{harmful conformity}, the probability of following the unanimous prior on trials where the prior is wrong.
\emph{Switch rate} (Experiment~2) is the proportion of items where the response differs from the no-prior baseline for the same model and item.
\emph{Switch direction} is assessed on switch trials only: $P(\text{switch}\rightarrow\text{human} \mid \text{switch})$, the proportion of switches that move toward the human source (as opposed to the opposing LLM). Because the two sources always disagree, values above 0.5 indicate a human-directed revision bias, and values below 0.5 indicate a revision biased toward the LLM source.

For Llama-3.3 70B we additionally extracted the log-probabilities assigned to the forced-choice answer tokens \texttt{YES} and \texttt{NO} at the final answer position\footnote{Log-probabilities were unavailable for other models via the API used in this study.}.
We normalize over \{\texttt{YES}, \texttt{NO}\} to obtain $p(\texttt{YES})$ and $p(\texttt{NO})$, and define $\mathrm{logit}(\texttt{YES})=\log p(\texttt{YES})-\log p(\texttt{NO})$.
In Experiment~1, we compute \emph{log-odds toward the prior} by re-coding the sign so that positive values indicate support for the prior answer (YES or NO).
In Experiment~2, we compute \emph{log-odds toward the human} analogously so that positive values indicate support for the human answer (YES or NO).
To isolate the effect of the social signal on token-level evidence, we compute within-item baseline-normalized shifts by subtracting the corresponding no-prior value for the same item (Experiment~1, $k{=}0$), i.e., $\Delta \mathrm{logit}=\mathrm{logit}_{\text{condition}}-\mathrm{logit}_{k=0}$.

\subsection{Statistical analysis}
For Experiment 1, we fit a logistic regression predicting conformity from source framing (Friends/Experts/Other LLMs), prior correctness, and group size (log-transformed), including interactions. To account for within-item dependence arising from repeated evaluation, we computed standard errors clustered by item (within each dataset and model). Dataset and model were included as fixed effects.
For Experiment 2, we fit logistic models predicting (i) switch likelihood and (ii) switch direction as a function of human source framing, again using item-clustered standard errors and controlling for dataset and model. All tests are two-sided unless otherwise noted.

\section{Results}

\subsection{Baseline performance}
Without priors ($k=0$), all models performed well on BoolQ and StrategyQA (BoolQ: 0.877--0.897; StrategyQA: 0.887--0.963), providing a strong baseline for evaluating the effect of social priors. Baseline performance on ETHICS was more variable across models (0.617--0.867), consistent with the greater ambiguity of normative judgments (Fig.~\ref{fig:exp1_all}).

\subsection{Experiment 1: Credibility-weighted influence under unanimous priors}
\begin{figure*}[ht!]
    \centering
    \includegraphics[width=\textwidth]{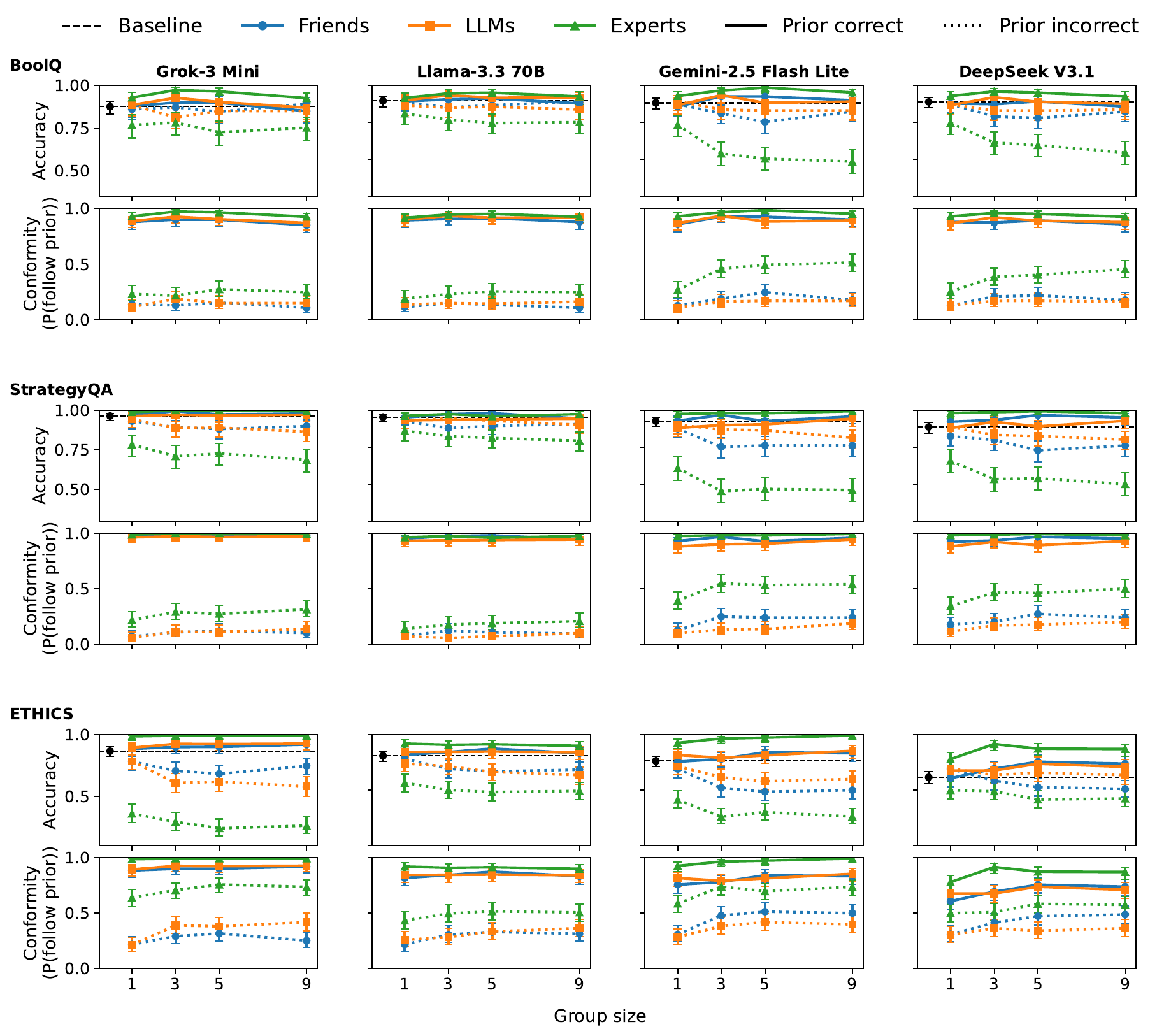}
    \caption{\textbf{Experiment 1: Effects of homogeneous social priors across datasets.}
    Rows correspond to datasets (BoolQ, StrategyQA, ETHICS) and columns to models.
    Within each dataset block, the top row shows accuracy and the bottom row shows conformity (probability of matching the unanimous prior) as a function of group size.
    Solid vs.\ dotted lines indicate whether the prior answer is correct or incorrect (agrees vs.\ disagrees with the dataset label).
    The dashed black line (and point at $k{=}0$) shows the no-prior baseline accuracy.
    Error bars show 95\% Wilson confidence intervals.}
    \label{fig:exp1_all}
\end{figure*}

Fig.~\ref{fig:exp1_all} shows results for all three datasets (BoolQ, StrategyQA, ETHICS; one row per dataset). For each source and group size $k$, we plot outcomes separately for trials where the unanimous prior is correct (solid lines) versus incorrect (dotted lines). Since priors were constructed to be correct with probability 0.5, the prior is uninformative about correctness on average; thus, systematic differences in conformity reflect how models weight the social signal based on its framing and strength.

Across all datasets, expert-framed priors exert substantially stronger influence than priors attributed to friends or other LLMs, and this influence increases with group size $k$ (Fig.~\ref{fig:exp1_all}). We quantify harmful influence as \emph{harmful conformity}, the probability of following an incorrect prior. At $k{=}9$, harmful conformity was consistently highest for experts: BoolQ (Experts 0.365 vs LLMs 0.160; risk difference RD = 0.205), StrategyQA (0.390 vs 0.155; RD = 0.235), and ETHICS (0.639 vs 0.387; RD = 0.252). In contrast, friend-framed priors were similar to other-LLM priors at $k{=}9$ (RD range across datasets: $-0.02$ to $0.01$), indicating that the large effects are specific to expert framing rather than the mere presence of social information.

Logistic regression confirmed these effects. Conformity increased sharply when the correct prior ($\beta{=}3.46 \pm 0.09$, $p{<}10^{-10}$) and increased with group size ($\beta{=}0.155 \pm 0.028$, $p{<}10^{-10}$); note that, given the above-chance baseline accuracy, it expected that conformity is higher when the randomized prior is correct. Expert framing substantially increased conformity relative to other-LLM priors ($\beta{=}1.00 \pm 0.06$, $p{<}10^{-10}$), whereas friend framing did not differ reliably from other LLMs ($\beta{=}0.066 \pm 0.059$, $p{=}0.27$). Together, these results indicate a credibility-weighted influence mechanism in which expert labels amplify both beneficial and harmful conformity, with larger unanimous groups yielding stronger effects.

\begin{figure}[t]
  \centering
  \begin{minipage}[t]{0.49\columnwidth}
    \textbf{(a)}\\[0.2ex]
    \includegraphics[width=\columnwidth]{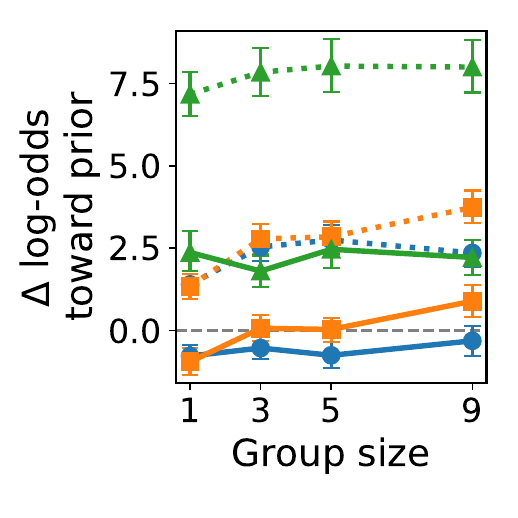}
  \end{minipage}\hfill
  \begin{minipage}[t]{0.49\columnwidth}
    \textbf{(b)}\\[0.2ex]
    \includegraphics[width=\columnwidth]{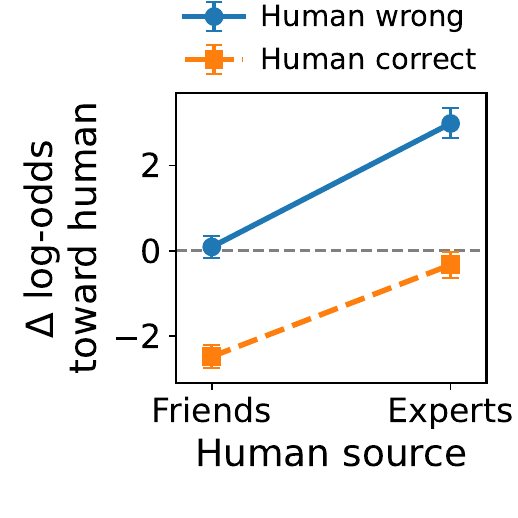}
  \end{minipage}

  \caption{\textbf{Token-level belief shifts (Llama-3.3 70B; BoolQ).}
\textbf{(a)} Experiment~1: Mean change in log-odds toward the unanimous prior (relative to no-prior baseline, $k{=}0$) as a function of group size.
Colors and line styles match Fig.~\ref{fig:exp1_all}.
\textbf{(b)} Experiment~2: Mean shift toward the human source under conflict ($k{=}2$) relative to baseline. 
Error bars show 95\% bootstrap confidence intervals.}

  \label{fig:logprob_shifts}
\end{figure}

To probe credibility framing at the level of decision evidence, we analyzed token-level \texttt{YES}/\texttt{NO} log-probabilities for Llama-3.3 70B on BoolQ.
Unanimous priors shift token-level evidence toward the stated answer, and this effect is strongly credibility-weighted: on trials where the prior is incorrect, expert framing produces much larger shifts than other-LLM or friend framing, and these shifts grow with group size (Fig.~\ref{fig:logprob_shifts}a).
At $k{=}9$, the log-odds shift toward an opposing prior is $7.97$ for experts (95\% bootstrap CI [$7.18$, $8.80$]), compared to $3.70$ for other-LLMs ([$3.18$, $4.21$]) and $2.31$ for friends ([$1.89$, $2.72$]).
These token-level results mirror the behavioral conformity patterns in Fig.~\ref{fig:exp1_all}, indicating that expert labels amplify the weight placed on the social signal at the level of decision evidence.

\subsection{Experiment 2: Belief revision under human--LLM conflict}
\begin{figure*}[ht!]
    \centering
    \includegraphics[width=\textwidth]{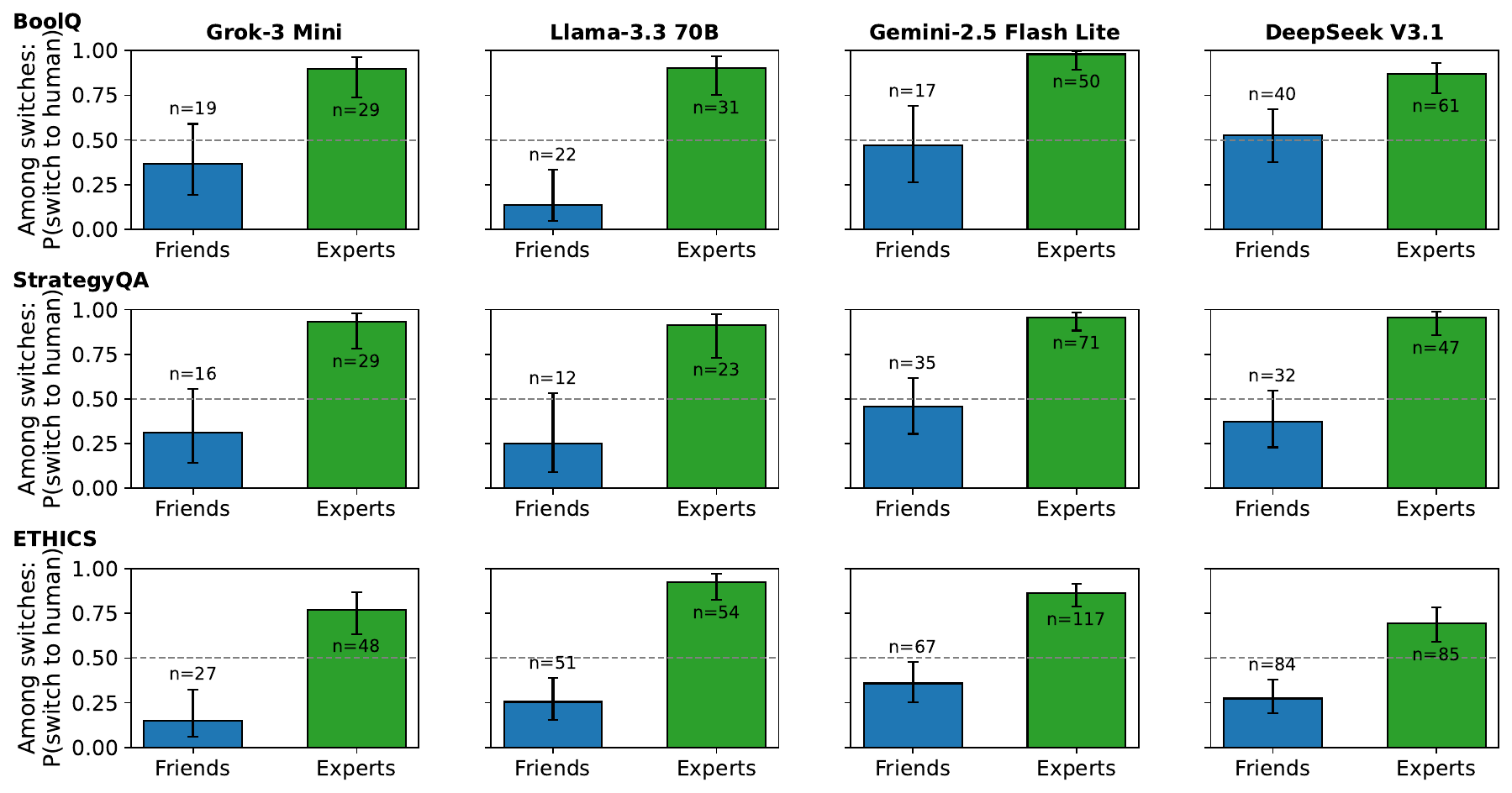}
    \caption{\textbf{Experiment 2: Belief revision under human--LLM conflict across datasets.}
    Rows correspond to datasets and columns to models.
    Considering only trials where the model’s final answer differs from its no-prior baseline (Experiment 1, $k{=}0$), bars show the probability that the switch follows the human source (Friends vs.\ Experts) rather than the opposing LLM.
    The dashed horizontal line marks 0.5 (no directional preference).
    Annotations report the number of switch trials ($n$) in each condition.
    Error bars show 95\% Wilson confidence intervals.}
    \label{fig:exp2_switchbias_all}
\end{figure*}

Given that expert framing produces the strongest influence in Experiment~1, including elevated conformity on disagreement trials, Experiment~2 tests whether the same credibility bias governs belief revision when a human and another LLM provide directly conflicting advice. To isolate belief revision, we analyze only trials where the model's final answer differs from its no-prior baseline (Experiment~1, $k=0$), and ask whether revisions move toward the human or toward the opposing LLM (Fig.~\ref{fig:exp2_switchbias_all}). Because the two sources always disagree, a revision toward one source implies a revision away from the other.

Across datasets, revisions are strongly biased toward expert-framed humans. Pooling across the four models within each dataset, the probability of switching toward the human expert (among switch trials) was 0.912 on BoolQ, 0.947 on StrategyQA, and 0.813 on ETHICS (all exact binomial tests vs.\ 0.5: $p{<}10^{-10}$). In contrast, revisions under friend framing were below 0.5 in all datasets (BoolQ: 0.398, StrategyQA: 0.379, ETHICS: 0.279), indicating that when models changed their mind under friend framing, they tended to move toward the opposing LLM rather than the friend (StrategyQA and ETHICS: $p<0.05$; BoolQ: one-sided $p=0.027$). Switching itself was relatively infrequent but occurred more often under expert framing than friend framing (switch rates, pooled across models: BoolQ 7.12\% vs 4.08\%; StrategyQA 7.08\% vs 3.96\%; ETHICS 12.67\% vs 9.54\%).

Logistic regression on switch trials confirmed a large expert-framing effect on revision direction: expert framing increased the odds of switching toward the human by a factor of 14.45 relative to friend framing (95\% CI [10.99, 19.02], $p{<}10^{-10}$), controlling for dataset and model with standard errors clustered by item.

Token-level analysis confirms this credibility hierarchy (Fig.~\ref{fig:logprob_shifts}b). Under conflict, expert framing shifted evidence strongly toward the human, even when the expert was wrong ($\Delta \mathrm{logit}_{\text{human}} = 2.98$, 95\% CI [$2.63$, $3.35$]).
In contrast, friend framing produced negligible shifts when the friend was wrong ($0.09$), and significant negative shifts (away from the human) when the friend was correct ($-2.48$). This indicates that the model implicitly tracks expert sources as high-value evidence while treating friends as less reliable than the opposing LLM.

\section{Discussion}

Our results demonstrate that LLM judgments are strongly shaped by social priors, with a pronounced credibility-weighting bias toward human experts. Across three datasets (BoolQ, StrategyQA, ETHICS) and two experimental paradigms, the central result is a robust credibility-weighted influence effect: identical priors produce much stronger shifts when attributed to human experts than when attributed to friends or other LLMs. Since our priors were uninformative on average, the observed shifts reflect pure sensitivity to source framing rather than objective reliability. This framing effect is bidirectional: expert priors amplify beneficial influence when the prior agrees with the dataset label and harmful influence when it disagrees, with larger unanimous groups yielding stronger effects. In direct human--LLM conflict (Experiment~2), belief revision is strongly biased toward expert-framed humans; under friend framing, revisions are instead biased toward the opposing LLM, as indicated by $P(\text{switch}\rightarrow\text{human}\mid\text{switch})<0.5$ across datasets.

\subsection{Relation to human conformity and social influence}
In human psychology, conformity reflects both informational and normative influence \citep{deutsch1955study}. Classic work shows that unanimity and group size increase conformity \citep{asch1951effects,asch1955opinions,bond2005group}. The present results mirror these determinants in one respect: increasing group size $k$ strengthens the impact of the social signal. However, the dominant determinant is not group structure per se but the credibility label attached to the group. This pattern aligns with decades of persuasion research showing that expert sources are weighted more heavily than low-credibility sources \citep{hovland1951influence}. Unlike human normative conformity, LLMs do not face social evaluation or interpersonal costs in the experimental setting. This suggests that the effect is best interpreted as a learned heuristic that treats expert framing as strong evidence and/or a learned tendency to comply with contextual cues about deference to expertise. 

\subsection{Connections to alignment, sycophancy, and LLM conformity}
A plausible mechanism is that instruction tuning and preference optimization reward cooperative behavior, including deference to contextual information, which may generalize to deference toward socially framed priors \citep{ouyang2022training}. Related work on sycophancy shows that RLHF-style assistants sometimes prioritize agreement with the user’s stated beliefs over truthfulness \citep{sharma2023towards}. While conformity to a group prior is distinct from user-directed sycophancy, both phenomena implicate sensitivity to social/credibility cues that can degrade epistemic reliability when the cue is wrong.

Recent work has reported conformity-like effects in LLMs and highlighted moderators such as uncertainty and interaction framing \citep{zhu2025conformity}. In multi-agent settings, conformity can reduce independence across agents and increase correlated errors, motivating benchmarks and mitigation strategies for groupthink-like dynamics \citep{weng2025do}. The present experiments complement this literature by isolating a particularly strong and general driver: \emph{who} is claimed to hold the belief (human experts vs friends vs other LLMs) can outweigh the informational content of the signal, and this effect replicates across factual QA (BoolQ), implicit reasoning (StrategyQA), and normative judgments (ETHICS). 

\subsection{Limitations and scope}
Several limitations qualify the interpretation of these effects. First, the priors are presented as short textual summaries rather than being generated by real humans or real expert panels; the experiments therefore measure sensitivity to source framing rather than calibrated learning about source reliability. Second, the design randomizes agreement with the dataset label on each item, which cleanly dissociates framing from true reliability but does not capture repeated interactions where reliability can be learned over time. Third, the tasks are forced-choice, which enables clear measurement of conformity and belief revision but may not generalize to open-ended generation, where models can hedge, qualify, or ask for clarification. Fourth, ETHICS labels serve as a benchmark target but represent normative judgments; while the expert-framing effect generalizes to this domain, ``correctness'' should be interpreted accordingly. Finally, the magnitude of effects may depend on prompt wording and training differences across model families; future work should assess robustness to alternative phrasing and richer social signals (e.g., rationales, confidence, reputational history), and should evaluate designs where sources have stable and learnable reliability.

\printbibliography


\end{document}